# Handwritten Character Recognition of South Indian Scripts: A Review

Jomy John, Pramod K. V, Kannan Balakrishnan

Department of Computer Applications

Cochin University of Science and Technology

Kochi 682 022

jomyeldos@gmail.com, pramod_k_v@cusat.ac.in, bkannan@cusat.ac.in

**Abstract:** *Handwritten character recognition is always a frontier area of research in the field of pattern recognition and image processing and there is a large demand for OCR on hand written documents. Even though, sufficient studies have performed in foreign scripts like Chinese, Japanese and Arabic characters, only a very few work can be traced for handwritten character recognition of Indian scripts especially for the South Indian scripts. This paper provides an overview of offline handwritten character recognition in South Indian Scripts, namely Malayalam, Tamil, Kannada and Telungu.*

## 1. INTRODUCTION

Handwritten character recognition is a frontier area of research for the past few decades and there is a large demand for OCR on handwritten documents. Even though, sufficient studies have performed in foreign scripts like Chinese, Japanese and Arabic characters [1], only a very few work can be traced for handwritten character recognition of Indian scripts. Even now no complete hand written text recognition system is available in Indian scenario and it is difficult due to large character set of Indian languages and the presence of vowel modifiers and compound characters in Indian script. Some reports have appeared for isolated handwritten characters and numerals of a few Indian languages. Majority of them was based on Bangla and Devnagiri script [2]. Nowadays, Technology Development for Indian languages (TDIL) and Resource for Indian language technology solutions (RCILTS), Ministry of Communication and Information Technology Solutions, Government of India are taken initiation towards development of language technology. Commercial systems are developed for some Indian scripts namely Assamese, Bangla, Devnagiri, Malayalam, Oriya, Tamil and Telungu, but that can handle only printed text, not handwritten manuscript

This study focuses mainly on offline handwritten character recognition of South Indian languages, namely, Telugu, Tamil, Malayalam and Kannada. Organisation of the paper is as follows. In section 2, we discuss about Indian language characteristics. Section 3 deals with the architecture of a general character recognition system with details of preprocessing, feature extraction and classification. Studies on handwritten characters of Malayalam, Tamil, Kannada and Telugu are covered in section 4, 5, 6 and 7 respectively. Section 8 concludes the paper.

## 2. INDIAN LANGUAGE CHARACTERISTICS

India is a multi lingual multi script country with twenty two scheduled languages, namely, Assamese, Bengali, Bodo, Dogri, Gujarati, Hindi, Kannada, Kashmiri, Konkani, Maithili, Malayalam, Manipuri (Meithei), Marathi, Nepali, Oriya, Punjabi, Sanskrit, Santali, Sindhi, Tamil, Telugu and Urdu [3]. These languages are written using only twelve scripts. Devnagiri script used to write Hindi, Konkani, Marathi, Nepali, Sanskrit, Bodo, Dogri and Mathili. Sindhi is written using Devnagiri script in India and Urdu script in Pakistan. Assamese, Manipuri and Bangla languages are written using Bengali script. Gurmukhi script is used to write Punjabi language. All other languages have their own script. In Indian language scripts, the concept of upper case and lower-case characters is not present. Most of the Indian languages are derived from Ancient Brahmi and are phonetic in nature and hence writing maps sounds of alphabets to specific shapes. All these languages, except Urdu, are written from left to right. The basic characters comprises of vowels and consonants. Two or more basic characters are combined to form compound characters

## 3. ARCHITECURE OF A GENERAL CHARACTER RECOGNITION SYSTEM

The major steps involved in recognition of characters include, pre processing, segmentation, feature extraction and classification (fig. 1)

### 3.1 PRE PROCESSING

The sequences of pre-processing steps are as follows

#### 3.1.1 *Noise Removal*

Noise is defined as any degradation in the image due to external disturbance. Quality of handwritten documents depends on various factors including quality of paper, aging of documents, quality of pen, color of ink etc. Some examples of noise are salt and pepper noise, Gaussian noise. These noises can be removed to certain extent using filtering technique. Technical details of filtering can be found in [4].

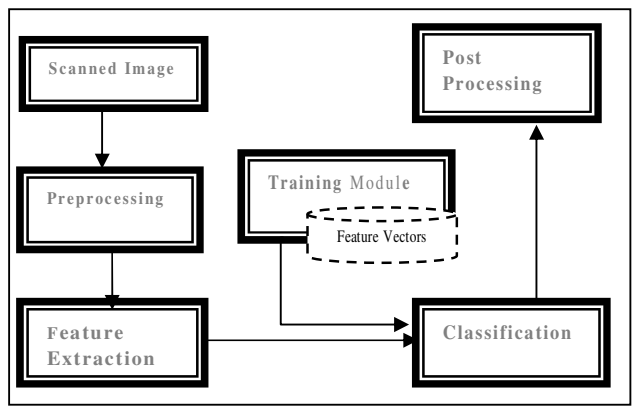

Figure 1.   Architecture of a character recogniton system

### 3.1.2    Thresholding

The task of thresholding is to extract the foreground (ink) from the background (paper) [5].  Given a threshold, T between 0 and 255, replace all the pixels with gray level lower than or equal to T with black (0), the rest with white (1). If the threshold is too low, it may reduce the number of objects and some objects may not be visible. If it is too high, we may include unwanted background information. The appropriate threshold value chosen can be applied globally or locally. Otsu's [6] algorithm is the commonly used global thresholding algorithm.

### 3.1.3    Skeletonization

Skeltonization is an image preprocessing operation performed to make the image crisper by reducing the binary-valued image regions to lines that approximate the skeletons of the region. A comprehensive survey of thinning methodologies is discussed in [7]

### 3.2    SEGMENTATION

Segmentation step contains line segmentation, word segmentation and character segmentation. Methods for character segmentations [8] are based on i) white space and pitch ii) projection analysis and iii) connected component labeling

### 3.3    NORMALIZATION

It is the process of converting the random sized image into standard sized image. This size normalization avoids inter class variation among characters. Bilinear, Bicubic interpolation techniques are a few methods for size normalization

### 3.4    FEATURE EXTRACTION

Features are a set of numbers that capture the salient characteristics of the segmented image.  There is different feature extraction methods proposed for character recognition [9].

### 3.5    CLASSIFICATION

The feature vector obtained from previous phase is assigned a class label and recognized using supervised and unsupervised method.  The data set is divided into training set and test set for each character. Character classifier can be Bayes classifier, Nearest neighbour classifier, Radial basis function, Support vector machine, Linear discriminant functions and Neural networks with or without back propagation

### 4.    STUDIES ON MALAYALAM HANDWRITTEN CHARACTER RECOGNITION

Malayalam is one of the four major Dravidian languages of South India and one among the twenty two scheduled languages of India with official language status in the State of Kerala and Union territories of Lakshadweep and Mahe, spoken by around 35 million people and ranked eighth in terms of the number of speakers. Malayalam script is derived from the Grantha script, an inheritor of olden Brahmi script. It is in close propinquity to Tamil and has indelible impression of Sanskrit. It also has the influence of Arabic. It is syllabic in nature and alphabets are classified into vowels and consonants. Conjunct symbols are used to combine certain consonants. At present 15 vowels (Fig 2) and 36 consonants are in use.

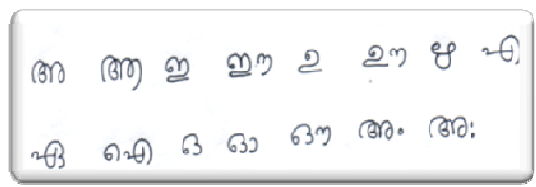

Figure 2.   Malayalam vowels

Lajish [10] in his work reported that it was the first work in handwritten Malayalam character recognition, in which fuzzy-zoned normalized vector distance features are classified using class modular neural network considering 44 Malayalam characters. Accuracy reported was 78.87%.

In another work [11], State space Point Distribution (SSPD) parameters derived from gray scale based SSM of handwritten character samples are utilized to obtain an accuracy of 73.03%.

Remarkable works on the application of Daubechie wavelet coefficients in HCR were reported by G. Raju [12]. In this work he used db4, a member of Daubechie wavelet family with order 4, for decomposition into ten sub-images (three levels decomposition using zero-crossing of wavelet coefficients for the recognition of unconstrained handwritten Malayalam characters. In [13], performance analysis of wavelet feature using twelve different wavelet filters was used for that study. An MLP network is used as classifier. It is observed that the performance of different wavelet filters used in the study is more or less same. The average recognition accuracy is 76.8%. The above work is extended by adding one more feature, the aspect ratio and found significant improvement in recognition, the average being 81.3%. In [14] count of zero-crossings in each of the sixteen sub bands together with a structural feature forms the feature vector. Feed forward back propagation network is used for classification with 90% accuracy in classification and recognition

R John [15] proposed a work using 1D wavelet transform of projection profiles as features. The preprocessed character images are modeled with projection profile. One dimensional wavelet transformation is applied on the projection profile. The feature vector is formed from the smooth components of the transform coefficients. A Multi Level Perceptron network is used for classification.

Recognition based on intensity pattern of characters was proposed by Rahiman [16].

Ref [17] deals with the recognition of handwritten Malayalam characters using discrete features. The features are extracted from skeletonizsed images. The skeleton pruning is done by contour portioning with discrete curve evolution with a recognition accuracy of 90.18 percent for 33 classes. In [18], Canny edge detector is used to produce thinned edges of the character and broken parts of edges are linked using ant colony optimization method. This image is further partitioned into different zones for the purpose of feature extraction. Multi layer perceptron (MLP) used these features and classified the characters with a recognition accuracy of 95.16%.

We have also developed a method that uses chain code and image centroid for the purpose of extracting features and a two layer feed forward network with scaled conjugate gradient for classification [19]. In [20], another approach for retrieval of similar Malayalam characters is described.

## 5. STUDIES ON TAMIL HANDWRITTEN CHARACTER RECOGNITION

Tamil is one of the oldest languages in India. It is the official language of the Indian state of Tamil Nadu and the union territories of Pondicherry and the Andaman and Nicobar Islands. It also has official status in Sri Lanka, Malaysia and Singapore. The Tamil script has 10 numerals, 12 vowels (fig. 3), 18 consonants and five grantha letters. The script, however, is syllabic and not alphabetic. The complete script, therefore, consists of 31 letters in their independent form, and an additional 216 combining letters representing every possible combination of a vowel and a consonant.

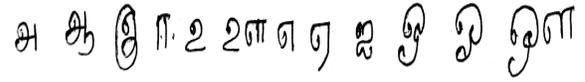

Figure 3.  Tamil vowels

A method to recognize hand printed Tamil characters was proposed in [21]. In this paper, authors have curves and strokes of characters. The input image is converted to labeled graph and correlation coefficients are computed for recognition of the character.

A hierarchical neural network which can recognize handwritten Tamil characters independently of their position and size are described by Paulpadian [22] in the year 1993.

In [23] pixel densities are calculated for different zones of the image and these values are used as the features of a character. These features are used to train and test the support vector machine. The results are tested for 3 different standard sizes of 32X32, 48X48 and 64X64. Recognition accuracy reported was 87.4%.

R. M. Suresh [24] attempts to use the fuzzy concept on handwritten Tamil characters to classify them as one among the prototype characters using distance from the frame and a suitable membership function. The prototype characters are categorized into two classes: one is considered as line characters/patterns and the other is arc patterns. The unknown input character is classified into one of these two classes first and then recognized to be one of the characters in that class. The algorithm is tested for about 250 samples for seven chosen Tamil characters and the success rate obtained varies from 88% to 100%.

Spatial space detection technique is used in [25], in which paragraphs are segmented into lines using vertical histogram, lines into words using horizontal histogram, and words into character image glyphs using horizontal histogram. The extracted features considered for recognition are given to Neural Network for classification.

Bhattacharya [26] proposed a recognition system for Tamil characters using the database of HP Lab India. The system used a two stage recognition approach. The training samples are first grouped using K-means clustering using a count of transition from one pixel position into other.  During the second stage MLP is used to classify each group using chain code histogram features of samples.  The recognition accuracy obtained is 92.77% and 89.66% for training and testing sets.

## 6. STUDIES ON KANNADA HANDWRITTEN CHARACTER RECOGNITION

Kannada is the official language of Karnataka and    is spoken by about 44 million people. The Kannada alphabets were developed from the Kadamba and Calukya scripts, descendents of Brahmi. The script has 49 characters in its alpha syllabic and is phonetic. There are 13 Vowels (Swara), 2 part vowel, part consonants (Yogavaha) and 34 Consonants

(Vangana) The script also includes 10 different Kannada numerals. Vowels are shown in fig.4

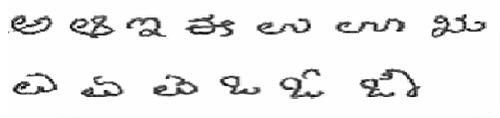

Figure 4.  Kannada vowels

A system for recognition of handwritten Kannada vowels by extracting invariant moments feature from zoned images is proposed in [27]. A Euclidian distance criterion and K-NN classifier is used to classify the handwritten Kannada vowels. A total 1625 images are considered for experimentation and overall accuracy found to be 85.53%.

Another method for recognition of printed and handwritten mixed Kannada numerals is presented using multi-class SVM for recognition yielding a recognition accuracy of 97.76% [28].

Rajashekararadhya [29] presented a Zone and Distance metric based feature extraction method for the recognition of Kannada and Telugu numerals. Feed forward back propagation neural network is used for classification and recognition with a recognition rate of 98% for isolated handwritten Kannada numerals and 96% for isolated handwritten Telugu numerals

Manjunath [30] reported a work on handwritten digit recognition based on radon transform. Radon function represents an image as a collection of projections along various directions. A Nearest neighbor classifier is used recognition purpose. The test was performed on the MNIST handwritten numeral database.

In [31], moments features are extracted from the Gabor wavelets of preprocessed images of 49 characters. The comparison of moments features of 4 directional images with original images when tested on Multi Layer Perceptron with Back Propagation Neural Network. The average performance of the system with these two features together is 92%.

Niranjan [32] proposed an unconstrained handwritten Kannada character recognition system based on Fisher Linear Discriminant Analysis (FLD). The proposed system extracts features from well known FLD, Two dimensional FLD (2D-FLD) and Diagonal FLD. In order to classify the characters, different distance measure techniques are used.

Ref [33], a handwritten Kannada and English Character recognition system based on spatial features is presented. Directional spatial features viz stroke density, stroke length and the number of stokes are employed as potential features to characterize the handwritten Kannada numerals/vowels and English uppercase alphabets. KNN classifier is used to classify the characters based on these features with four fold cross validation. The proposed system achieves the recognition accuracy as 96.2%, 90.1% and 91.04% for handwritten

Kannada numerals, vowels and English uppercase alphabets respectively

7.  STUDIES ON TELUNGU HANDWRITTEN CHARACTER RECOGNITION

Telugu is the Dravidian language and it is the third most popular scripts in India. It is the official language of the southern Indian state, Andhra Pradesh and also spoken by neighboring states. Telugu is a syllabic language. The Telugu scripts are closely related to the Kannada script. Officially, there are 10 numerals, 18 vowels (fig.5), 36 consonants and three dual symbols

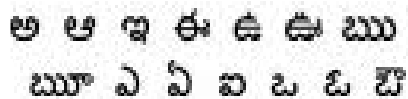

Figure 5.  Telungu vowels

In 2008, Rajashekararadhya [34] proposed an offline handwritten numeral recognition technique for four south Indian languages like Kannada, Telugu, Tamil and Malayalam. In this work they suggested a feature extraction technique, based on zone and image centroid. They used two different classifiers nearest neighbor and back propagation neural network to achieve 99% accuracy for Kannada and Telugu, 96% for Tamil and 95% for Malayalam

In [35] Kannada, Telugu and Devnagari handwritten numerals are considered recognition system using global and local structural features and a Probabilistic Neural Network (PNN) classifier. The feature set include directional density estimation, water reservoir method, filled hole density and maximum profile distance. The average recognition rate of 99.40%, 99.60% and 98.40 obtained for Kannada, Telugu and Devnagari datasets respectively. It is reported to be thinning free, and without size normalization

Pal [36] proposed a quadratic classifier based scheme for the recognition of off-line handwritten characters of three popular south Indian scripts: Kannada, Telugu, and Tamil. The features used are mainly obtained from the directional information. For feature computation, the bounding box of a character is segmented into blocks, and the directional features are computed in each block. These blocks are then down-sampled by a Gaussian filter, and the features obtained from the down-sampled blocks are fed to a modified quadratic classifier for recognition. They used 64-dimensional features for high speed recognition and 400-dimensional features for high accuracy recognition. A five-fold cross validation technique was used for result computation, and obtained 90.34%, 90.90%, and 96.73% accuracy rates from Kannada, Telugu, and Tamil characters, respectively, from 400 dimensional features. Pal [37] proposed another work using a modified quadratic classifier towards the recognition of off-line handwritten numerals of six popular Indian scripts Devnagari, Bangla, Telugu, Oriya, Kannada, and Tamil.

## 8. CONCLUSION

In this paper, character recognition systems for handwritten Malayalam, Tamil, Telungu and Kannada script are discussed in detail. Different segmentation techniques and various classifiers with different features are also discussed. Notwithstanding the importance and the need, this problem is not adequately investigated by the researchers. One of the major difficulties in this field is the lack of bench mark database for hand written characters for most of the languages for testing of research results. We believe that our survey will be helpful for researchers in this field.